\documentclass[10pt,twocolumn,letterpaper]{article}

\usepackage{cvpr}
\usepackage{times}
\usepackage{epsfig}
\usepackage{graphicx}
\usepackage{amsmath}
\usepackage{amssymb}
\usepackage{multirow}
\usepackage{rotating}
\usepackage{comment}

\usepackage[pagebackref=true,breaklinks=true,letterpaper=true,colorlinks,bookmarks=false]{hyperref}

\cvprfinalcopy 

\def\cvprPaperID{5156} 

\begin{document}

\title{A Structured Model For Action Detection}

\author{Yubo Zhang, Pavel Tokmakov, Martial Hebert\\
Carnegie Mellon University\\
{\tt\small yuboz,ptokmako,hebert@andrew.cmu.edu}
\and
Cordelia Schmid\\
Google Research\\
{\tt\small cordelias@google.com}
}

\maketitle
\thispagestyle{empty}

\begin{abstract}
A dominant paradigm for learning-based approaches in computer vision is training generic models, such as ResNet for image recognition, or I3D for video understanding, on large datasets and allowing them to discover the optimal representation for the problem at hand. While this is an obviously attractive approach, it is not applicable in all scenarios. We claim that action detection is one such challenging problem - the models that need to be trained are large, and the labeled data is expensive to obtain. To address this limitation, we propose to incorporate domain knowledge into the structure of the model to simplify optimization. In particular, we augment a standard I3D network with a tracking module to aggregate long term motion patterns, and use a graph convolutional network to reason about interactions between actors and objects. Evaluated on the challenging AVA dataset, the proposed approach improves over the I3D baseline by 5.5\% mAP and over the state-of-the-art by 4.8\% mAP. 
\end{abstract}

\vspace{-5mm}

\section{Introduction}

\begin{figure}[t]
    \centering
    \includegraphics[width=0.94\linewidth]{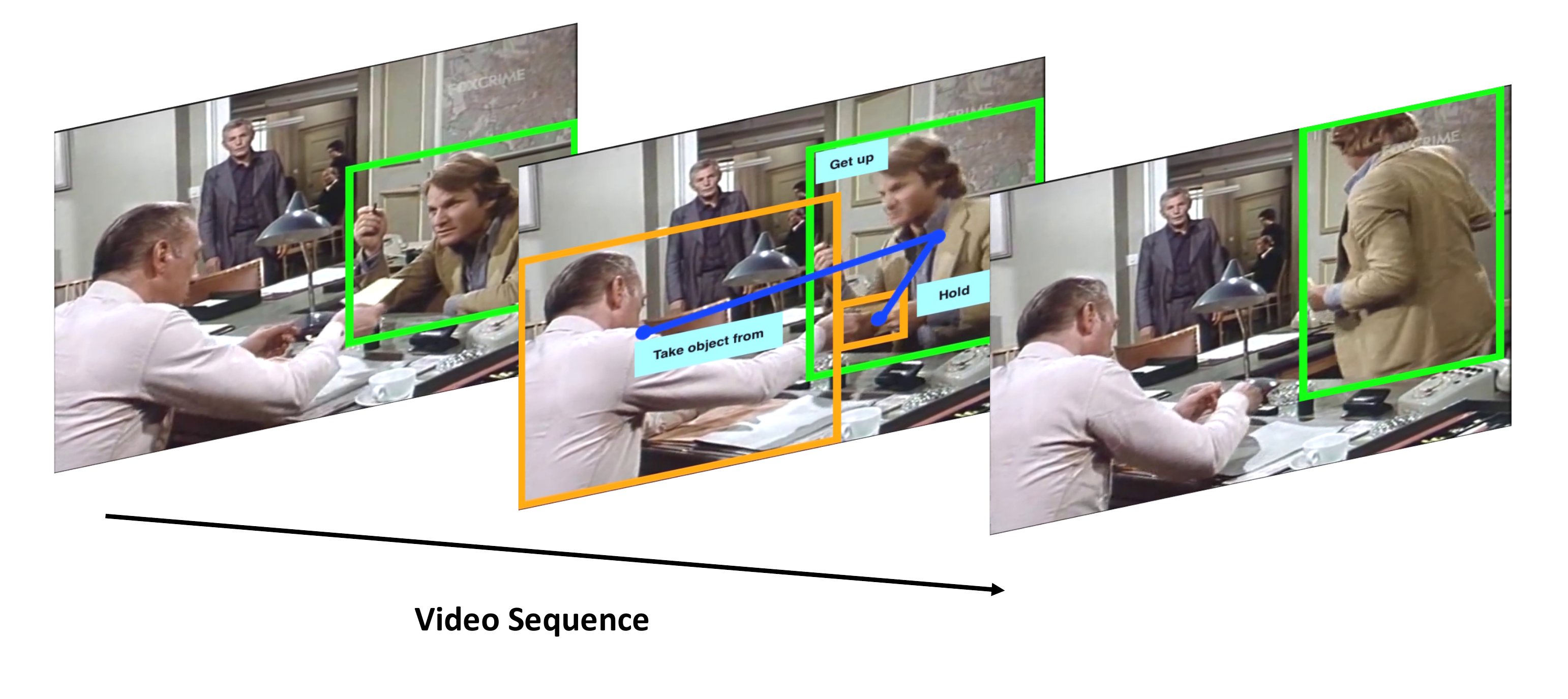}
    \vspace{-3mm}
    \caption{For action detection, it is critical to capture both the long-term temporal information and spatial relationships between actors and objects. We propose to incorporate this domain knowledge into the architecture of deep learning models for action detection. }
    \vspace{-5mm}
    \label{fig:teaser}
\end{figure}

Consider the video sequence from the AVA dataset~\cite{gu2017ava} shown in Figure~\ref{fig:teaser}. It shows a person getting up and then receiving a letter from another person, who is seated behind a table. Out of the 2359296 pixels in the 36 frames of this clip, what information is actually important for recognizing and localizing this action? Key cues include the location of the actor, his motion, and his interactions with the other actor and the letter. The rest of the video content, such as the color of the walls or the lamp on the table are irrelevant and should be marginalized over. We use these intuitive observations to design a new method for action detection. 


State-of-the-art action detection approaches put a lot of emphasis on actor localization~\cite{gu2017ava,hou2017tube,kalogeiton2017action,singh2017online}, but other cues are largely ignored. For instance, Gu et al.~\cite{gu2017ava} detect humans and model their actions with an I3D~\cite{carreira2017quo} representation that is capable of capturing short-term motion patterns. This allows them to achieve a significant improvement on the challenging AVA dataset,  but the performance on activities with large temporal extent remains poor. In our method, we aggregate local I3D features over actor tracks, which results in a significant gain in performance.

A few recent approaches model human-object interaction. 
Gkioxari et al.~\cite{gkioxari2017detecting} use a state-of-the-art 2D-object detection framework~\cite{he2017mask} to detect action specific objects and  model human-object interactions in static images. Their approach assumes the object categories 
given and does not integrate any temporal information. 
Sun et al.~\cite{sun2018actor} addressed the problem of modeling human-human and human-object interaction, by applying relational networks to explicitly capture interactions between actors and objects in a scene. Their method, however, does not directly model objects, but instead considers every pixel in the frame to be an object proxy. While this approach is indeed generic and object-category agnostic, we argue that the lack of proper object modeling hinders its performance. In a concurrent work to~\cite{sun2018actor}, Wang et al.~\cite{wang2018videos} use object proposals to localize the regions of interest and then employ graph convolutional networks~\cite{kipf2016semi} to combine the actor and object representations and produce video-level action classification. However, their approach does not address the action detection problem. In our method we also model activities with actor-object graphs, but instead of aggregating features over all the objects and actors in a scene we propose to structurally modeling actor-object and actor-actor separately during both training and testing. Other works that propose to capture action recognition with actor-object graphs include~\cite{ibrahim2016hierarchical,qi2018learning}. These methods, however, require ground truth annotations of both actors and objects during training and focus on a closed vocabulary of object categories. Our method addresses both of these limitations by first adopting a weakly-supervised object detection approach for localizing the correct objects during training time without explicit supervision, and secondly proposing a simple modification to the state-of-the-art object detection framework~\cite{he2017mask} which makes it category agnostic.

In this work we propose a model for action detection in videos that explicitly models long-term human behaviour, as well as human-human and human-object interactions. In particular, our model extracts I3D~\cite{carreira2017quo} features for the frames in a video sequence and, in parallel, detects persons and objects with an object detection approach modified from He et al.~\cite{he2017mask} (Sec \ref{sec:frontend}). It then tracks every actor over a 3-second interval producing a set of \textit{tubelets}, e.g. sequences of bounding boxes over time~\cite{kalogeiton2017action,kang2016object}. To this end a simple and efficient heuristic tracker is proposed (Sec \ref{sec:association}). The tubelets are then combined with the detected objects to construct an actor-centric graph (Sec \ref{sec:graph}). Features from an I3D frame encoding are pooled to obtain a representation for the nodes. Every edge in the graph captures a possible human-human or human-object interaction. A classifier is then trained on the edge features to produce the final predictions. Naively, such an approach requires ground truth object annotation to train. To remove this requirement we build on intuition from weakly-supervised object detection and learn to integrate useful information from the objects at training time automatically.


To summarize, this work has two main contributions: (1) We propose a new method for action detection that explicitly captures long-term behaviour as well as human-human and human-object interactions; (2) We demonstration state-of-the-art results on the challenging AVA dataset, improving over the best published method by 4.8\%, and provide a comprehensive ablative analysis of our approach.

\section{Related work}

\noindent \textbf{Action classification} is one of the fundamental problems in computer vision. Early approaches relied on hand-crafted features~\cite{wang2013dense} that track pixels over time and then aggregated their motion statistics into compact video descriptors. With the arrival of deep learning these methods have been outperformed by two-stream networks~\cite{simonyan2014two} that take both raw images and optical flow fields as input to CNNs~\cite{lecun1995convolutional}, which are trained end-to-end on large datasets. These methods are limited by the 2D nature of CNN representations. This limitation has been addressed by Tran et al.~\cite{tran2015learning} who extended CNN filters to the temporal dimension resulting in 3D convolutional networks. More recently, Carreira and Zisserman~\cite{carreira2017quo} have integrated 3D convolutions into a state-of-the-art 2D CNN architecture~\cite{szegedy2015going}, resulting in Inflated 3D ConvNet (I3D). Wang et al.~\cite{wang2017non}, have extended this architecture with non-local blocks that facilitate fine-grained action recognition. We use an I3D with non-local blocks as the video feature representation in our model.

\noindent \textbf{Action localization} can refer to spatial, temporal, or spatio-temporal localization of actions in videos. In this work we study the problem of spatial action localization. Early action detection methods~\cite{klaser2010human,prest2013explicit} generate hand-crafted features from videos and train SVM classifier. Early deep-learning based action localization models~\cite{Gkioxari_2015_CVPR,peng2016multi,saha2016deep,singh2017online,Weinzaepfel_2015_ICCV} are developed on top of 2D object detection architectures. They detect actors in every frame and recognize activities using 2D appearance features.
Kalogeiton et al.~\cite{kalogeiton2017action} proposed to predict short tubelets instead of boxes by taking several frames as input. However their model only uses tubelets for temporal localization. In Li et al.~\cite{li2018recurrent} the authors apply an LSTM~\cite{gers1999learning} on top of the tubelet features to exploit long-term temporal information for action detection. However, their model also relies on a 2D representation and is not trained end-to-end. TCNN~\cite{hou2017tube} uses C3D as a feature representation for action localization, but they only extract features for a single bounding box in the middle of a short sequence of frames. Finally, Gu et al.~\cite{gu2017ava} propose to use I3D as a feature representation, which takes longer video sequences as input, but also does not aggregate the features over a tubelet. Our model builds upon the success of I3D for feature extraction. 
Instead of extracting I3D features for the entire video given a single location, we track actors based on their appearance and extract their feature representations along the entire video clip, which enables learning discriminate features for actions with long temporal dependency.

\begin{figure*}[t]
\vspace{-8mm}
    \centering
    \includegraphics[width=1.0\linewidth]{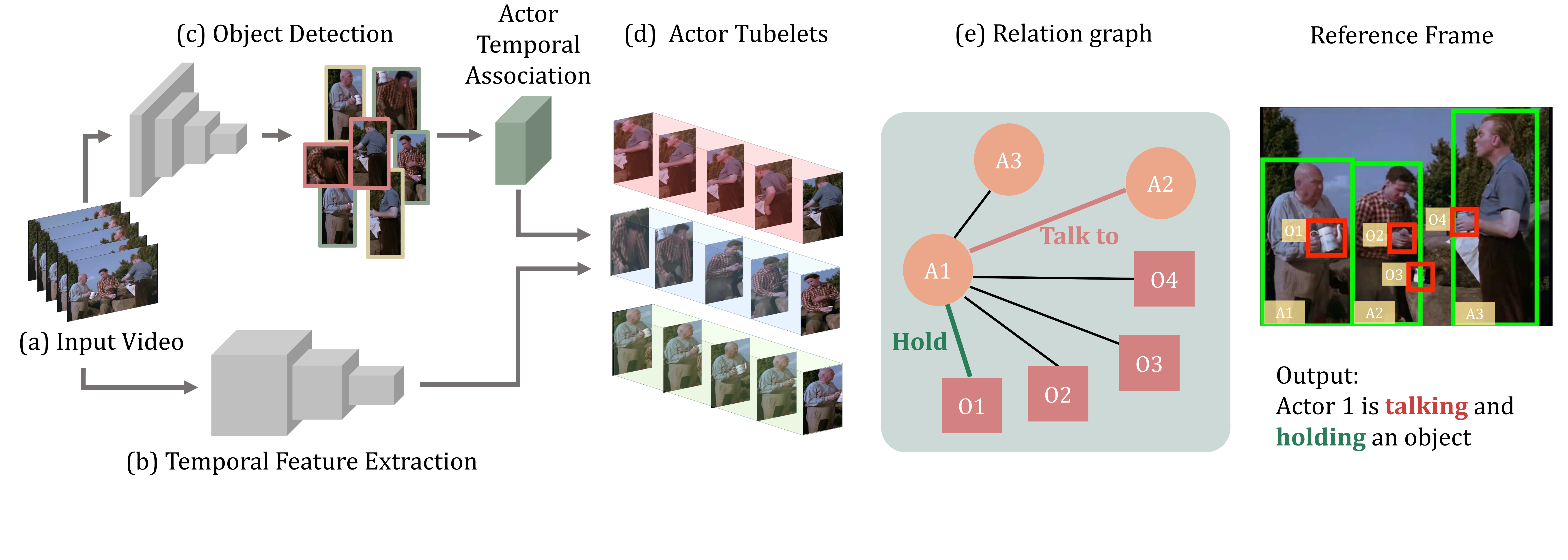}
    \vspace{-12mm}
    \caption{Overview of our proposed framework. We model both long-term person behaviour and human-human, human-object interactions structurally in a unified framework. The actors across the video are associated to generate actor tubelets for learning long temporal dependency. The features from actor tubelets and object proposals are then used to construct a relation graph to model human-object manipulation and human-human interaction actions. The output of our model are actor-centric actions.}
    \vspace{-4mm}
    \label{fig:model}
\end{figure*}

\noindent \textbf{Object detection} is a key component of most of the action detection frameworks. Traditional approaches relied on hand-crafted features and part-based models~\cite{felzenszwalb2010object}. Modern deep-learning based methods are either based on RCNN-like~\cite{Girshick_2015_ICCV,girshick14CVPR,he2017mask,ren2015faster}, or SSD-like architectures~\cite{liu2016ssd,redmon2016you}. In our model, we use Mask-RCNN~\cite{he2017mask} for person and object detection. 
To detect any objects that participate in interactions we employ the method of Dave et al.~\cite{dave2019segmenting}, who propose a simple modification of the training procedure of Mask-RCNN, making the model category-agnostic.

\noindent \textbf{Object tracking} is a well studied problem. Traditional tracking algorithms~\cite{baker2004lucas,henriques2015high,kalal2012tracking} used hand-crafted appearance features to perform online tracking of the bounding box in the first frame. Despite their efficiency, the performance of these methods on realistic videos is sub-optimal. State-of-the-art, deep learning-based trackers~\cite{feichtenhofer2017detect,hong2015online,ma2015hierarchical,tao2016siamese,zhu2016robust} demonstrate a better performance and are more robust. 
Our tracking module, following the tracking by detection paradigm, first detects all humans in consecutive video frames. Instead of online fine-tuning the model on the detected actors in the first frame, we propose to train a siamese-network~\cite{bertinetto2016fully} offline with a triplet loss. 

\textbf{Visual relationship modeling} for human-human and human-object pairs increases performance in a variety of tasks including action recognition~\cite{wang2018videos} and image captioning~\cite{ma2017attend,peyre2017weakly}. There have been several works~\cite{chao2018learning,gkioxari2017detecting,gupta2015visual} on human-object interaction modeling in images that achieved significant improvements on HICO-DET~\cite{chao2015hico} and V-COCO~\cite{lin2014microsoft} datasets. Kalogeiton et al.~\cite{kalogeiton2017joint} train object and action detection models together and jointly predict object-action pairs. Their model requires all annotation of objects and only uses 2D CNNs. Mettes et al.~\cite{mettes2017spatial} encode the features from actors, objects and their spatial relation into a single representation to model actions for zero-shot learning. Recently, Qi et al. \cite{qi2018learning} propose a framework for action localization in videos which represents humans, objects and their interactions with a graphical model. It then uses convolutional LSTMs~\cite{xingjian2015convolutional} to model the evolution of the graph over time. Their model, however, uses 2D CNNs for feature representation, requires ground truth annotations of the object boxes for training and is only evaluated on a toy dataset~\cite{koppula2016anticipating}. Baradel et al.~\cite{baradel2018object} propose to use object relation network to model the temporal evolution of objects for action recognition. However, their method also relies on object class annotation and they are not modeling the relationship between the objects and the actors. Our model does not require object annotations which allows us to demonstrate results in a more realistic scenario. Similarly to us, Sun et al.~\cite{sun2018actor} propose to implicitly model the interactions between actors and objects without object annotations for training. To this end they use relational networks~\cite{santoro2017simple} which avoid explicitly modeling objects by treating each location in an image as an object proxy and aggregating the representations across all the locations. In our evaluation we show that explicit modeling of objects and integration of the relevant objects in a frame allows us to learn more discriminative features.


\section{Method}

We propose a method for action detection in videos that explicitly models the long-term behaviour of individual people, along with human-human and human-object interactions. The architecture of our model is shown in Figure~\ref{fig:model}. It takes a sequence of video frames as input (a) and passes them through an I3D network (b). In parallel, a state-of-the-art object detection model~\cite{he2017mask} (c) is applied to each frame to produce human and object bounding boxes. Human bounding boxes are then combined into \textit{tubelets} (a sequence of bounding boxes over time) (d) with an association module. The tubelets and object boxes (as nodes) are then used to construct an \textit{actor-centric graph} for every actor in the video clip (e). 

In the actor-centric graph, we define two kinds of nodes, the \textit{actor node} and the \textit{object node}, along with two kinds of edges, representing human-object manipulation and human-human interaction. The \textit{object nodes} are generated by performing Region of Interest (ROI) Pooling from the I3D representation. The \textit{actor nodes}, whose temporal behavior we wish to model, are obtained by aggregating I3D features with graph convolutions over the corresponding tubelets. The features from the graph edges are used as the final representation for action classification. The whole model, except for the 2D object detector, is trained in an end-to-end fashion requiring only actor bounding boxes and ground truth actions. In the rest of this section, we will first present our models for video representation and object detection. Then, we explain how we integrate temporal information using an appearance-based multi-object tracking module. Finally, we will demonstrate how we build the actor-centric graph, and how it is used to generate action predictions.

\subsection{Spatio-temporal feature extraction} \label{sec:frontend}
The first step in our action detection pipeline is to extract two sets of features from videos: an unstructured video embedding, and a collection of object and actor region proposals. 

\noindent \textbf{Unstructured video embedding.} To exploit the spatio-temporal structure of the video input, we use an inflated 3D ConvNet (I3D) with non-local layers~\cite{wang2018videos}. In a 3D ConvNet, videos are modeled as a dense sampling of $x,y,t$ coordinates, and the corresponding learned filters operate in both spatial and temporal domains, thus capturing short-term motion patterns. We also use non-local layers~\cite{wang2017non} to aggregate features across the entire image, allowing our network to reason beyond the extend of local convolutional filters. In our scenario, the input is a 3 seconds video clip with 36 frames. Our final video embedding retains its temporal dimension, enabling us to explicitly use temporal information in the later stages of our model.

\noindent \textbf{Appearance based actors/objects proposal.} We take advantage of the success of RCNN-like models~\cite{ren2015faster}  for object detection to identify regions of interest. In our model, we are interested in identifying the spatial location of the actors and potential objects that are being manipulated by them. Since our goal is to understand actions performed by humans, independent of the categories of objects, we use a category-agnostic detector proposed in~\cite{dave2019segmenting} to localize the objects. This model achieves a higher recall for the objects that are not among the 80 categories labeled in MS-COCO. Specifically, we train Mask-RCNN~\cite{he2017mask} on MS-COCO \cite{lin2014microsoft} by collapsing all the category labels into a single {\tt object} label, resulting in a category-agnostic object detector. We use a standard person detector for localizing the actors~\cite{he2017mask}.

\subsection{Action detection with temporal context}

To enable our action detection system to capture long-term temporal dependencies, we integrate multi-object tracking into our action detection framework. Instead of generating explicit action proposals, we track each actor across frames in the entire video. Then, with the actor appearance information stored in a node and tracking information in edges, we aggregate each actor's movement by using graph convolutions.

\subsubsection{Multi-actor association module} \label{sec:association}

We note that some actions are composed of multiple unit movements, for example, the action 'get up' is composed of siting, moving upward, and standing. We posit that confidently tracking actors across multiple frames and integrating these local representations in a principled way is crucial for learning discriminative representations for actions that are composed of multiple movements. Previous methods that recognize actions from a few frames and link them via actioness score ~\cite{singh2017online} are not able to maintain consistent tracks, since, unlike the appearance features, the features of a model trained for action recognition differ significantly across frames due to the actor's movement.

Motivated by this observation, we introduce a multi-actor association module that aims to associate the bounding box proposals of each actor throughout the video clip. Instead of linking action bounding box proposals based on actionness scores, we associate actor bounding boxes based on the similarity of actor appearance features. 

 We follow the tracking-by-detection paradigm, and build an association module to perform the linking. Specifically, we first train an appearance feature encoding, and then explicitly search over neighbor regions in the next frame for an appearance match. To learn an appearance feature encoding for distinguishing different actors, we train a Siamese network~\cite{hermans2017defense} with a triplet loss~\cite{schroff2015facenet}. After we obtain the appearance feature encoding, we search among the bounding box proposals in consecutive frames and match the bounding boxes with highest appearance similarity.

\subsubsection{Actor tubelet learning using graphs} \label{sec:graph}
Recent works in action detection attempt to predict an action directly from the features extracted from I3D~\cite{gu2017ava}. We claim that integrating I3D features over multiple frames is crucial for recognizing long-term activities. A naive approach would be to simply average these features along the temporal dimension. Instead we propose to model the behavior of each actor with graph convolutional networks~\cite{kipf2016semi}. We propose to encode the nodes of the person graph with features extracted from an I3D backbone with RoIAlign~\cite{he2017mask}. The edges are obtained from the tubelets constructed by our multi-actor association module. While performing graph convolutions, the movement information of each actor box is aggregated by the graph. Formally, let us assume that there are $N$ actors in a video. Each actor is represented by a feature vector of dimension $D$. $T$ is the temporal dimension. We denote by $G$ the affinity matrix of the actor tubelet graph with dimension $N \times T$, and by $X$ the actor features with dimension $T \times D$. The graph convolution operation can the be written as $Y = GXW $, where $W$ is the matrix of weights with dimension $D \times D$. The output of the graph $Y$ has the dimension $N \times D$ and aggregates the actors' features along the temporal axis. The graph convolution operations can also be stacked in multiple layers to learn more discriminative features.

%
\subsection{Interactions between actors and objects}
To recognize actions associated with interactions, it is critical to exploit the relations between the actor of interest, other actors, and objects in the scene. However, modeling all such possible relationships can become intractable. We propose to use class-agnostic features from ROI proposals to build a relation graph and implicitly perform relation reasoning given only action annotations. 

To integrate information from the other actors and objects, we construct two relation graphs, one to model human-object manipulation and the other one to model human-human interaction. The human-object graph connects each actor of interest with the other objects and the human-human graph connects each actor of interest with the other actors. The features of actor nodes come from the actor tubelets after the multi-actor association module and we denote them with $H = [h_1,h_2,...,h_N]$ where $N$ is the number of actors in the middle frame of a clip. The features of the objects are generated by ROI pooling of I3D representation and are denoted as $O = [o_1,o_2,...o_M]$ where $M$ is the number of objects in the whole video. 

To model relationships between a selected actor and other subjects, we can build on the concepts of \textit{hard} and \textit{soft} attention models~\cite{xu2015show}. One way to represent the features of the actions is to first localize the correct subjects among all the objects and all the other actors (except the target actor). Then, one can use the features from the actor and the identified subjects, which we refer as \textit{hard relation graph}. Alternatively, in the \textit{soft relation graph}, instead of explicitly localizing the subjects, we integrate this information by implicitly learning how much they relate to the target actor. We will further demonstrate how we implement soft relation graph and hard relation graph to learn discriminative feature representation for interactions.

\noindent \textbf{Hard relation graph.} We explicitly localize the correct objects and actors for each target actor to represent the object manipulation actions and human interaction actions. The object manipulation action is represented through linking an actor node and the object nodes, while the human interaction action is represented through the edges between one actor and the other actor nodes. Given actor node features $H = [h_1,h_2,...,h_N]$ and object node features $O = [o_1,o_2,...o_M]$, the object-manipulation relation feature for the $i^{th}$ target actor and the $j^{th}$ object can be represented by concatenating the features of the two nodes with 
\begin{equation}
    f_{h_i, o_j} = F_o([h_i, o_j]),
\end{equation}
where $F_o$ is the feature extraction function for object manipulation. Similarly, with $F_h$ being the feature extraction function for human interaction, we represent the human interaction relation feature for the $i^{th}$ and the $k^{th}$ actor with
\begin{equation}
    f_{h_i, h_k} = F_h([h_i, h_k]),
\end{equation}

In the absence of ground truth annotations for the target objects, we resort to an approach inspired by multi-instance learning for object detection, and select the region with the maximal score for the ground truth action. Specifically, for an object manipulation action centered at the $i^{th}$ actor,
\begin{equation}
    \hat{p^{i}_o} = \max_j \sigma(f_{h_i, o_j}),
\end{equation}

where $\sigma$ is the sigmoid function, and $\hat{p^{i}_o}$ is the human-object manipulation action prediction for the $i^th$ actor. Similarly, the prediction for human interaction actions is
\begin{equation}
    \hat{p^{i}_h} = \max_k \sigma(f_{h_i, h_k}),
\end{equation}
where $\hat{p^{i}_h}$ is the human-human interaction action prediction for the $i^th$ actor.

\noindent \textbf{Soft relation graph.} The hard approach described above is appealing conceptually, but results in instability during training. We thus propose an alternative method that avoids making hard decisions about the ground truth objects by aggregating the information over all the objects in the scene. We define the strength of a relation between the actor of interest and another actor or object as the inverse of Euclidean distance between the two nodes' features after a feature transformation.

The transformations for actor features and object features are defined with with $\phi_h$ and $\phi_o$ respectively. Given actor node features $H = [h_1,h_2,...,h_N]$ and object node features $O = [o_1,o_2,...o_M]$, we first transform them to obtain $\phi_h(H) = [\phi_h(h_1),\phi_h(h_2),...,\phi_h(h_N)]$, $\phi_o(O) = [\phi_o(o_1),\phi_o(o_2),...,\phi_o(o_N)]$. The edge between the $i^{th}$ actor and the $j^{th}$ object is represented with 
\begin{equation}
    f_o(h_i,o_j) = \frac{1}{\lVert \phi_h(h_i) - \phi_o(o_j)\rVert_2}.
\end{equation}
The edge between the $i^{th}$ actor and the $k^{th}$ actor is represented similarly.

We further normalize the edge weights above so that they sum to one. We adopt softmax function for each actor with

\begin{equation}
    G^o_{ij} = \frac{ \exp f_o(h_i,o_j)}{\sum^{M}_{m=1}\exp f_o(h_i,o_m)},
\end{equation}

\begin{equation}
    G^h_{ik} = \frac{ \exp f_h(h_i,h_k)}{\sum^{N-1}_{n=1}\exp f_h(h_i,h_n)},
\end{equation}
where $k$ is 1...N except $i$.

After computing the graph representation, the object manipulation and human interaction actions for the $i^{th}$ actor are represented with

\begin{equation}
    F^o_{i} = \phi_h(h_i) + \sum^{M}_{j=1}G^o_{ij} \phi_o(o_j),
    \label{eq:human_feat}
\end{equation}

\begin{equation}
    F^h_{i} = \phi_h(h_i) + \sum^{N-1}_{k=1}G^h_{ik} \phi_h(h_k).
    \label{eq:object_feat}
\end{equation}
The final action predictions are obtained by logistic classifiers applied to the feature representation in the Equations~\ref{eq:human_feat} and~\ref{eq:object_feat} for human-human, and human-object interaction classes respectively.

\begin{figure*}[t]
\vspace{-6mm}
    \centering
    \includegraphics[width=1.0\linewidth]{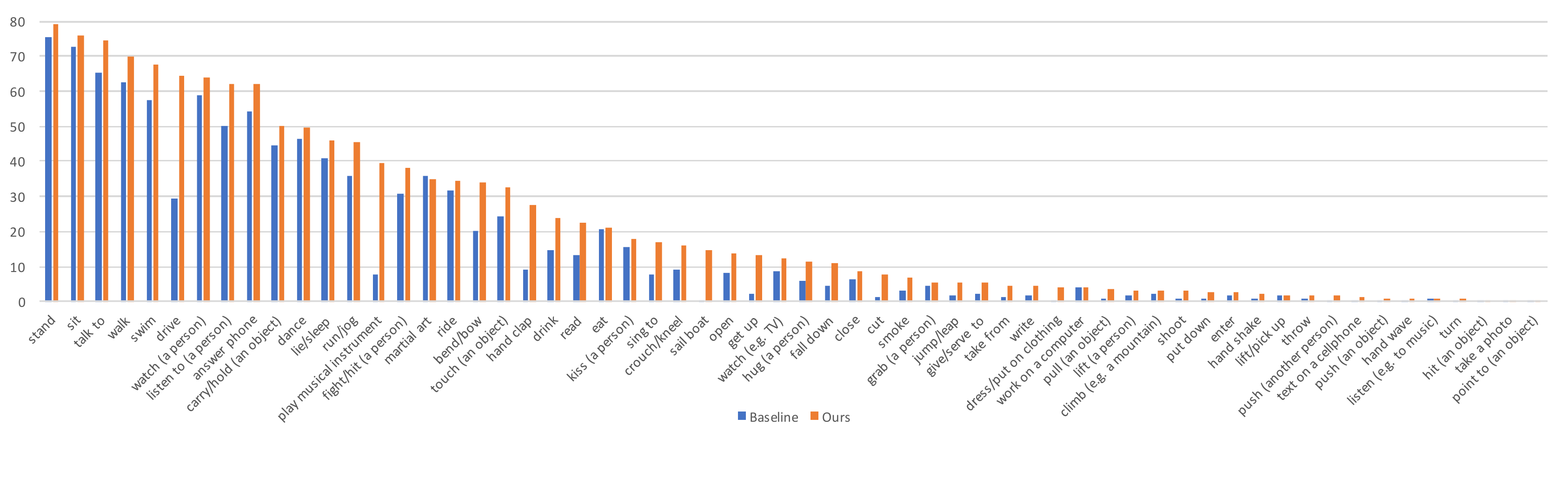}
    \vspace{-8mm}
    \caption{Per-category results for the proposed model and the baseline on the validation set of AVA.}
    \vspace{-5mm}
    \label{x}
\end{figure*}

\section{Experiments}

In this section, we first introduce the dataset and the metrics used for the evaluation of our model, and describe the implementation details. Next, we perform an extensive ablation analysis, demonstrating the effectiveness of our model on integrating temporal and spatial context information. Finally we compare our model with state-of-the-art methods both quantitatively and qualitatively.

\subsection{Datasets and metric}

We develop our model on the AVA version 2.1 benchmark dataset~\cite{gu2017ava}, where action localization is evaluated on the middle frame of three seconds videos clips. The video clips are extracted from movies and extensively annotated with bounding boxes of all the actors and the actions they are performing. Thus, this dataset is realistic both in terms of appearance and in terms of the label distribution. It contains 211k training samples and 57k validation samples. There are 80 categories in the dataset and 60 categories with no less than 25 validation samples are used for evaluation. We report frame based mean average precision with an intersection-over-union (IOU) threshold 0.5.

We also evaluate the performance of our model on the UCF-101 \cite{soomro2012ucf101} dataset. We report the results on split1 which contains 2293 training and 914 validation clips. There are 24 action categories. As in AVA, we report frame based mean average precision with an IOU threshold 0.5.

\subsection{Implementation details}

Our model is implemented in the Caffe2 framework. We follow the schema as proposed in~\cite{carreira2017quo, wang2017non} to pre-train our video backbone model. We use the ResNet-50 architecture and pretrain it on the ImageNet dataset~\cite{russakovsky2015imagenet}. The model is then inflated into 3D ConvNet as proposed in~\cite{carreira2017quo} (I3D), and pretrained on Kinetics dataset~\cite{carreira2017quo}. We augment our backbone model with non-local operations \cite{ wang2017non} after Res2, Res3, and Res4 blocks. We further fine tune it end-to-end with our proposed spatio-temporal model. Our video backbone model takes video clips of 36 frames as input corresponding to 3-second video clips at 12 fps. The frames are first scaled to 272 $\times$ 272, and randomly cropped to 256 $\times$ 256.

For region proposal model, we use Mask-RCNN~\cite{he2017mask} with a ResNet-50 backbone. 
We limit the set of labels to {\tt person} and {\tt object} only. The region proposal model is pretrained on COCO dataset~\cite{lin2014microsoft} and further fine tuned on AVA. We use 0.5 as threshold for object bounding boxes and 0.9 for person bounding boxes.

We trained our model on 8-GPU machine where each GPU has 3 video clips as mini-batch. The total batch size is 24. We freeze parameters in batch normalization layers during training and apply a drop out layer before the final layer. We use a drop out rate of 0.3. We first train for 90K iteration with learning rate 0.00125 and then train for another 10K iterations with learning rate 0.000125.

\begin{table}[t]
\centering 
\begin{tabular}{c |c} 
\hline\hline 
Model & mAP  \\ [0.5ex] 
\hline 
Baseline & 16.7  \\ 
Person similarity graph on ROIs & 20.1 \\
Object similarity graph on ROIs & 20.3 \\
Actor tubelets model & 21.1  \\
Actor tubelets + hard relation graph module & 21.5\\
Actor tubelets + soft relation graph module & 22.2\\
\hline
\end{tabular}
\vspace{2mm}
\caption{Analysis of different components of our model on the validation set of AVA.} 
\vspace{-5mm}
\label{table:evaluation} 
\end{table}

For the tracking module, we use a ResNet-50 architecture for appearance feature encoding and triplet loss~\cite{schroff2015facenet} to learn representative appearance features for tracking actors in the video. The model takes three images as input where two of them are the cropped images of the same actor at different time (ranging from 0.02s to 10s) and the third one is the cropped area of a different actor sampled from the same period. The output feature dimension is 128 and we use L2 distance as similarity metric. The model is fine tuned from ImageNet pretrained weights for 100K iterations with a batch size of 64. While tracking, we search over region of interest proposals with an overlap larger than 0.5 with the bounding box in the previous frame, and link the boxes which minimize the L2 distance in the embedding space. 

\begin{table*}[t]
\vspace{3mm}
\centering 
\begin{tabular}{c | c | c | c} 
\hline\hline 
Model & Human pose & Object manipulation & Human interaction  \\ [0.5ex] 
\hline 
Baseline & 35.7& 8.9&16.9  \\ 
Person similarity graph on ROIs &39.1 & 12.1& 20.1\\
Object similarity graph on ROIs & 39.3&13.0 &20.0 \\
Actor tubelets model & 40.6& 13.4& 20.9 \\
Actor tubelets + hard relation graph module &41.0 & 13.2&22.2 \\
Actor tubelets + soft relation graph module &41.9 & 14.3& 22.0 \\
\hline

\end{tabular}
\vspace{2mm}
\caption{Ablation analysis on human pose, human-object manipulation and human-human interaction categories. } 
\vspace{-5mm}
\label{table:per-category} 
\end{table*}

\subsection{Ablation analysis}

We first perform an ablation analysis of our framework to understand the effect of each component of the model 
in Table~\ref{table:evaluation}. We then perform a more in-depth analysis of the model by separately evaluating human pose, object manipulation, and human interaction classes in Table~\ref{table:per-category}.

All our models are developed on the non-local augmented I3D backbone. The baseline averages the I3D features over the temporal dimension, and uses actor bounding boxes to pool the features for action recognition. It achieves an mAP of 16.7 on the validation set, which is slightly improved compared to the baseline established in~\cite{sun2018actor}. 

\begin{figure*}[t]
\vspace{-9mm}
    \centering
    \includegraphics[width=1.0\linewidth]{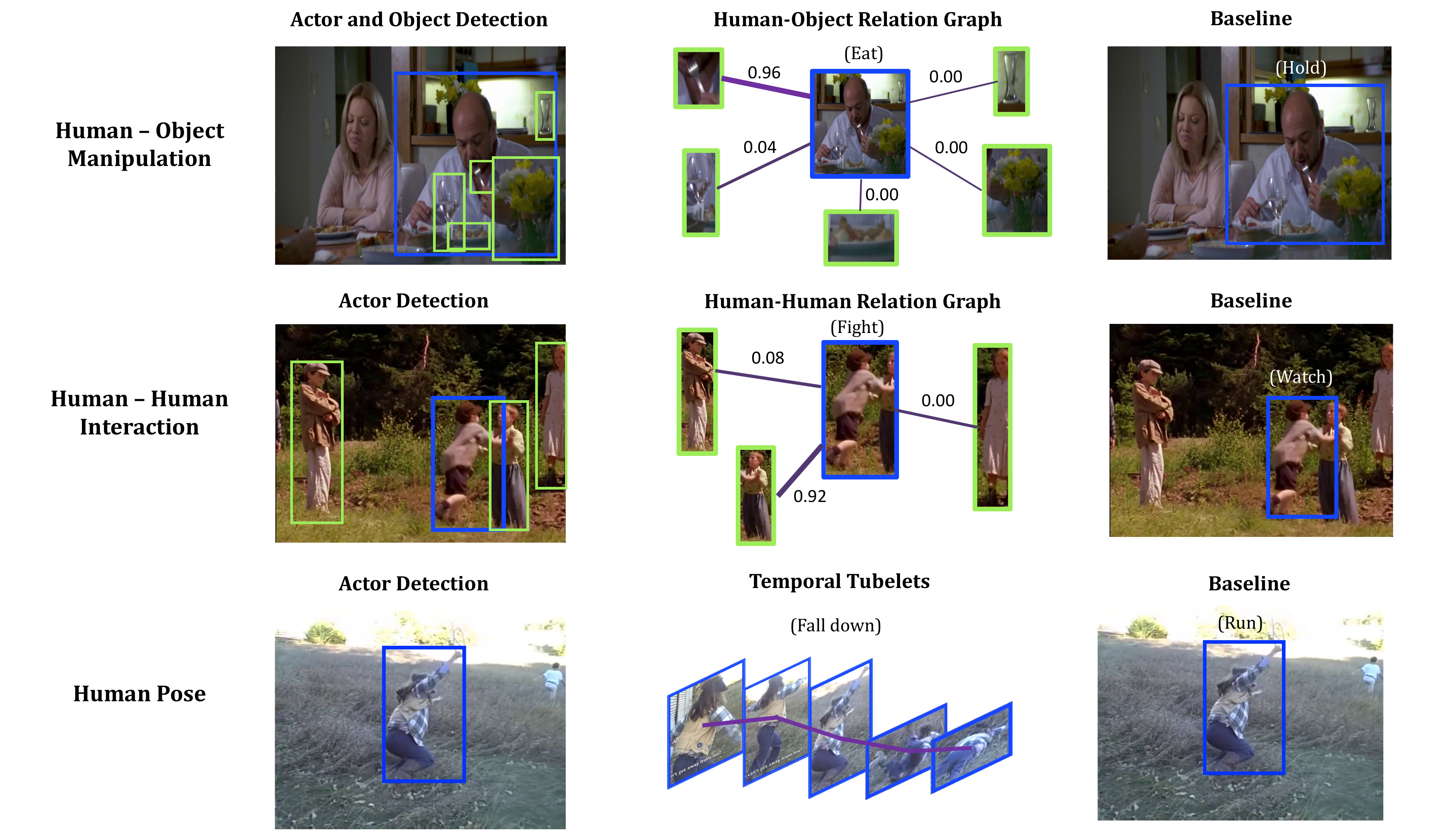}
    \vspace{-5mm}
    \caption{We visualize the performance of our model and the baseline. We show actor and object detections used by our model in the first column, the corresponding instantiations of the graphs in the second column, and baseline results in the third column.}
    
    \vspace{-5mm}
    \label{fig:vis}
\end{figure*}


We now introduce two additional baselines. Wang et al. \cite{wang2018videos} propose to use a similarity graph and a spatio-temporal graph to integrate information spatially and temporally for action recognition. We adapt their work to the domain of action detection, where actor proposals occur across the frames and the similarity graph integrates information over frames. We observe that the model that explicitly builds a similarity graph on all human proposals in the whole video achieves an mAP 20.1 on the validation set. As a second baseline, 
we build a similarity graph model over all the object proposals in the video clip. This model includes both humans and objects to provide information for modeling interactions, and achieves a score of 20.3 mAP. By integrating information from regions of interest spatially and temporally, both the person similarity graph and the object similarity graph achieve a significant increase over the baseline.

We now analyze different components of our approach. The actor tubelets model explicitly connects the same actor across frames and applies graph convolutions to aggregate the motion information. This basic variant, which does not model actor interactions achieves an mAP score of 21.1, which is a 4.4\% improvement over the baseline and 1\% improvement over the person similarity graph. Notice that both  approaches use person regions of interest. The better performance of actor tubelets model shows that explicitly tracking the actor helps our model to learn a better representation for action detection. Next we evaluate our hard relation and soft relation graph for learning actions involving interaction. The hard relation graph model achieves mAP 21.5 and the soft relation graph model achieves the best performance with mAP 22.2. This is probably due to the instability in training of the hard variant. The performance boost from our relation graph models further validates the efficiency of our proposed structured network architecture for modelling temporal dependencies and interactions. 

In addition to the averaged score over all 60 test classes, we also show performance on the three action categories: human pose, object manipulation and human interaction in Table~\ref{table:per-category}. We observe that our actor tubelet model largely outperforms the person graph model and the baseline on human pose categories and object manipulation categories. Further with soft relation graph, we observe that the mAP on human pose, object manipulation and human interaction action increases 6.2, 5.4 and 5.1 compared to the baseline respectively which demonstrates the effectiveness of our model for modeling both temporal dependency and interactions. We also visualize per-class mAP comparing our actor tubelet with soft relation graph model and the baseline in Figure~\ref{fig:model}. According to our observation, the largest improvement over the baseline is achieved on categories drive, play musical instrument and hand clap which are actions requiring learning long term temporal dependencies and capturing interactions with objects.

\subsection{Comparison to the state-of-the-art}

\begin{table}[t]
\centering 

\begin{tabular}{c |c} 
\hline\hline 
Model & mAP  \\ [0.5ex] 
\hline 
Single Frame model \cite{gu2017ava} & 14.2  \\ 
ACRN \cite{sun2018actor} & 17.4 \\
Our model & 22.2 \\
\hline
\end{tabular}
\vspace{1mm}
\caption{Comparison of our model to the state-of-the-art methods on the validation set of AVA.} 
\vspace{-5mm}
\label{table:state-of-the-art} 
\end{table}

In this section we compare our best model to the state-of-the-art models on the AVA dataset and on UCF-101-24 dataset ~\cite{soomro2012ucf101}. The performance on AVA is shown in the Table ~\ref{table:state-of-the-art}. Our proposed approach outperforms the method of Sun et al.,~\cite{sun2018actor} by 4.8\%. This is due to the inductive biases encoded into the architecture of our model via the actor tracking module, human-human and human-object relational graphs. In contrast, ACRN~\cite{sun2018actor} models relation by considering every pixel in the frame as an object proxy which is a less strong constraint. It is also not able to integrate long-term human motion information. 

We additionally evaluate our model on UCF-101-24 dataset, where our model with an actor tubelet and a human-object soft relation graph achieves an mAP score of 77.9, compared to 72.0 achieved by the baseline. We note that our model is still 0.9 mAP points bellow the state-of-the-art reported in~\cite{xie2018rethinking}. However, their model uses an S3D network as a backbone, which is shown to give a 6.8 mAP boost compared to the I3D. This suggests that our performance can be further improved by switching to a better backbone.


\subsection{Qualitative analysis}

In order to qualitatively evaluate our model, we verify its ability to capture temporal information and contextual relations. We visualize video clips and provide a performance comparison on several challenging examples in Figure~\ref{fig:vis}. In these examples actors are performing actions with nontrivial temporal behavior and challenging object interactions.

In the first row, we show a man eating with a fork. The baseline confuses the action with {\tt hold}, failing to incorporate information spatially from the dining table and the fork. Our human-object relational graph in contrast is able to aggregate this information efficiently. As shown in the third column, the edge between the person and the fork has a high value, which helps our model to make a correct prediction.

The second row shows two children who are fighting. The baseline mistakenly predicts the category {\tt watch}, since it does not integrate the features from both actors. Our model, however, use a human-human relation graph to reason about both actors jointly. As shown in the visualization of the graph, the edge between the key actor and the boy he is fighting with has a high value, which helps our model to correctly recognize the action.

In the third row, we show the action {\tt fall down}. To model this action, it is crucial to integrate information from both temporal and spatial domains as it is uniquely defined as a sequence of movements from standing to lying. Our model is able to correctly recognize this class by accumulating the temporal information with large spatial displacements. However, the baseline model mistakenly predicts the action as {\tt run}, since it only integrates features in a fixed bounding box area.



\vspace{-2mm}
\section{Conclusion}
\vspace{-1mm}
We proposed a structured model for action detection that explicitly models long-term temporal behavior as well as object manipulation and human interaction. Our model demonstrates large performance gains over the state-of-the-art, which highlights the effectiveness of our method in modeling temporal dependencies and reasoning about interactions. More importantly, the success of our model shows the importance of integrating temporal and relational information in the model architecture for the task of action detection.
\vspace{-4mm}
\paragraph{Acknowledgements.}

We thank Chieh-En Tsai, Mengtian Li, Leonid Keselman, Xiaolong Wang, Achal Dave for reviewing early versions of this paper and discussions. Supported by Google Cloud Platform. Supported by the Intelligence Advanced Research Projects Activity (IARPA) via Department of Interior/Interior Business Center (DOI/IBC) contract
number D17PC00345. The U.S. Government is authorized
to reproduce and distribute reprints for Governmental
purposes not withstanding any copyright annotation
theron. Disclaimer: The views and conclusions contained
herein are those of the authors and should not be interpreted
as necessarily representing the official policies or endorsements,
either expressed or implied of IARPA, DOI/IBC or
the U.S. Government.

{\small
\bibliographystyle{ieee}
\bibliography{egbib}
}

\end{document}